# Simple Modification of the Upper Confidence Bound Algorithm by Generalized Weighted Averages

**Nobuhito Manome**[1], **Shuji Shinohara**[1,2], **Ung-il Chung**[1]

[1] The University of Tokyo
[2] Tokyo Denki University
{manome, tei}@bioeng.t.u-tokyo.ac.jp, s.shinohara@mail.dendai.ac.jp

## Abstract

The multi-armed bandit (MAB) problem is a classical problem that models sequential decision-making under uncertainty in reinforcement learning. In this study, we propose a new generalized upper confidence bound (UCB) algorithm (GWA-UCB1) by extending UCB1, which is a representative algorithm for MAB problems, using generalized weighted averages, and present an effective algorithm for various problem settings. GWA-UCB1 is a two-parameter generalization of the balance between exploration and exploitation in UCB1 and can be implemented with a simple modification of the UCB1 formula. Therefore, this algorithm can be easily applied to UCB-based reinforcement learning models. In preliminary experiments, we investigated the optimal parameters of a simple generalized UCB1 (G-UCB1), prepared for comparison and GWA-UCB1, in a stochastic MAB problem with two arms. Subsequently, we confirmed the performance of the algorithms with the investigated parameters on stochastic MAB problems when arm reward probabilities were sampled from uniform or normal distributions and on survival MAB problems assuming more realistic situations. GWA-UCB1 outperformed G-UCB1, UCB1-Tuned, and Thompson sampling in most problem settings and can be useful in many situations. The code is available at https://github.com/manome/python-mab.

## Introduction

The multi-armed bandit (MAB) problem refers to the problem of maximizing gain in a setting with multiple arms, where a reward can be obtained from the arms with a certain probability by choosing an arm (Robbins 1952). This problem is considered the most fundamental in reinforcement learning because it involves a tradeoff between exploration for an arm with a high reward probability and exploitation to select an arm that is believed to have a high reward probability (Sutton and Barto 1998). The MAB problem has a wide range of applications, including online advertising (Xu et al. 2013; Schwartz et al. 2017; Nuara et al. 2018), recommendation systems (Elena et al. 2021; Silva et al. 2022), and games (Kocsis and Szepesvári 2006; Moraes et al. 2018; Świechowski et al. 2023). Algorithms for MAB problems are useful in many situations; however, it is often difficult to determine which algorithm to use. Typical algorithms for MAB problems include upper confidence bound (UCB) policies (Auer et al. 2002) and Thompson sampling (Thompson 1933). Thompson sampling achieves empirically superior performance (Chapelle and Li 2011) and is known to outperform UCB policies within a finite number of trials (Kaufmann et al. 2012b). These algorithms are popular because of their ease of implementation and guaranteed near-optimal theoretical performance (Bubeck and Liu 2013; Russo and Van Roy 2016). Therefore, both UCB policies and Thompson sampling are frequently applied in reinforcement learning (Auer et al. 2008; Osband et al. 2013). Two state-of-the-art algorithms, information-directed sampling (IDS) (Russo and Van Roy 2018) and TS-UCB (Baek and Farias 2023), are also based on them, but both require large amounts of computation.

In this study, we extend UCB1, the most representative algorithm among UCB policies that uses generalized weighted averages, and propose a new generalized UCB (GWA-UCB1) that can be executed with low computational cost. GWA-UCB1 is a two-parameter generalization of the balance of exploration and exploitation in UCB1 and can be implemented by simply modifying the UCB1 formula. UCB-based models such as Q-learning models with UCB (Jin et al. 2018) and reinforcement learning models with a Monte Carlo tree search (Silver et al. 2017; Silver et al. 2018; Schrittwieser et al. 2020) may be easily extended. The contribution of this study is the presentation of a set of parameters and an MAB algorithm that can be used effectively in various problem settings with low computational cost. However, it is important to note that it does not guarantee optimal regret from a theoretical perspective. The GWA-UCB1 with two parameters proposed in this study outperformed Thompson sampling in many situations. Furthermore, by adjusting the parameters according to the environment, performance close to that of the IDS was achieved.

The remainder of this paper is organized as follows. First, we introduce a review of UCB policies. Next, we explain the stochastic MAB problem and survival MAB problem, which are frameworks of MAB problems. Finally, we describe the proposed GWA-UCB1, discuss the experiments conducted to demonstrate its performance, and present the results and utility of the proposed method.

## Related Works

To maximize the reward obtained in the MAB problem, it is necessary to strike a balance between searching for the best arm and using knowledge obtained from the search results about the reward probability of the arm to select the arm believed to be the best. UCB policies are algorithms with a theoretically guaranteed upper bound on the expected loss (Auer et al. 2002), and they strike a good balance between exploration and exploitation. UCB1 is a representative algorithm; it first selects all arms at once and then selects the arm with the highest UCB score, as defined in Equation (1).

$$UCB_i^{UCB1} = \bar{X}_i + \sqrt{\frac{2 \log n}{T_i(n)}}, \quad (1)$$

where $\bar{X}_i$ is the expected value of arm $i$, $n$ is the number of selections for all arms, and $T_i(n)$ is the number of selections for arm $i$. Because the UCB score increases as $T_i(n)$ decreases, even if $\bar{X}_i$ is small, an arm with a small number of samples is inevitably more likely to be selected. The formula for UCB1 is based on the theoretical limits of Hoeffding's inequality.

UCB1-Tuned is an improved UCB1 model that considers the variance in the empirical value of each arm. The UCB score of the UCB1-Tuned is calculated using Equation (2).

$$UCB_i^{UCB1-Tuned} = \bar{X}_i + \sqrt{\frac{\log n}{T_i(n)} \min\left\{\frac{1}{4}, V_i(T_i(n))\right\}} \quad (2)$$

$V_i$ considers the variance of the reward and is calculated using Equation (3).

$$V_i(s) = \left(\frac{1}{s}\sum_{\gamma=1}^{s} X_{i,\gamma}^2\right) - \bar{X}_{i,s}^2 + \sqrt{\frac{2 \log n}{s}}, \quad (3)$$

where $X_{i,\gamma}$ is the reward for arm $i$ at time $\gamma$.

Another extension of UCB is Bayes-UCB, which constructs upper confidence bounds based on the quantiles of the posterior distribution (Kaufmann et al. 2012a; Kaufmann 2018). The formula for UCB1 is based on Hoeffding's inequality, but it can also be calculated using the Chernoff-Hoeffding inequality, which provides a more precise upper bound on the probability. It is known as KL-UCB because it is computed using the Kullback–Leibler (KL) divergence (Garivier and Cappé 2011; Cappé et al. 2013). Additionally, KL-UCB+ (Garivier et al. 2016) and KL-UCB++ (Ménard and Garivier 2017) have been proposed as improved KL-UCB models. These algorithms achieve high performance but are computationally intensive compared with UCB1. A recent study presented TS-UCB, which computes the scores for each arm using both posterior samples and confidence limits at each step (Baek and Farias 2023). In particular, TS-UCB guarantees optimal regrets and achieves a performance equal to or better than that of the state-of-the-art algorithm IDS (Russo and Van Roy 2018). However, IDS and TS-UCB also require a large amount of computation.

Although there are several generalizations of the UCB algorithm (Liu et al. 2019; Korkut and Li 2021), the simplest is the UCT formula (Kocsis and Szepesvári 2006), which applies the UCB algorithm to a Monte Carlo tree search. In this study, we call the generalized UCB1 (G-UCB1) a simple generalization of UCB1 with one parameter, similar to UCT, and the UCB score in G-UCB1 is calculated using Equation (4).

$$UCB_i^{G-UCB1} = \bar{X}_i + c\sqrt{\frac{2 \log n}{T_i(n)}}, \quad (4)$$

where $c$ is a constant for adjusting the weight of the exploration amount in the range $c \geq 0$. When $c$ is multiplied by $1/\sqrt{2}$, it matches the UCT formula. Increasing c means increasing the amount of search, which agrees with Equation (1) of UCB1 when $c = 1$. Equation (4) balances exploration and exploitation with a single parameter; however, our idea is to extend this balance to two parameters using generalized weighted averages. The details are explained in a subsequent section.

## Problem Setup

This section describes the stochastic MAB problem, which is the most common MAB problem, and the survival MAB problem, which involves the risk of ruins.

### Stochastic Multi-Armed Bandit Problem

The game flow for a typical MAB problem is as follows. Initially, the agent selects arm $i$ from a set of $k$ possible arms based on its strategy. Subsequently, the reward $r_i$ obtained by selecting arm $i$ is checked, and the arm to be selected next is sequentially determined.

The stochastic MAB problem is an MAB problem in which the rewards for arms are assumed to be generated according to a probability distribution, and the reward for arm $i$ is determined based on the reward probability $p_i$ set for each arm. In this study, the agent selects arm $i$ to receive reward 1 with probability $p_i$ and reward 0 with probability $1 - p_i$. The number of times an arm is selected is called a step, and an agent can select only one arm per step. The algorithm is evaluated using the expected regret, which is the difference between the expected reward of the arm actually selected and the arm with the highest reward probability in all steps.

**Survival Multi-Armed Bandit Problem**

The survival MAB problem is a recent extension of the MAB problem in which agents must maintain a positive budget throughout the process (Perotto et al. 2019). This problem is similar to the budgeted MAB problem (Tran-Thanh et al. 2010), but the true risk of ruin must be considered.

The game flow for the survival MAB problem is the same as that for the probabilistic MAB problem; however, the agent has an initial budget $b_0$ at the start of the game, and care must be taken to ensure that the budget reaches 0. In this study, if the agent chooses arm $i$, it receives reward 1 with probability $p_i$ and reward $-1$ with probability $1 - p_i$. In other words, the initial budget $b_0$ changes to $b_{t+1} = b_t + r_t$ depending on the observed reward. The algorithm is evaluated using the survival rate, which is the percentage that was not ruined in each step, and the budget.

Solving the survival MAB problem involves managing the cost of exploratory behavior to avoid ruin while maximizing the budget. Therefore, the dilemma between exploration and exploitation is more complex and difficult to analyze than the usual MAB problem; however, it is a more practical problem setting. For example, stock traders have finite amounts of money and must avoid bankruptcy. All organisms in nature must avoid death while maximizing important variables specific to the organism. The purpose of this study was not to theoretically optimize regret but to present a heuristic algorithm that can be used effectively in many problem settings. Therefore, no theoretical analysis was performed in this study, but there is a study (Riou et al. 2022) that presents a Pareto optimal policy for the probability of ruin in the survival MAB problem; we refer the reader to this paper for theoretical considerations.

## Proposed Method

As mentioned previously, a balance between exploration and exploitation is important for maximizing the rewards obtained in the MAB problem. Averages are often used to balance two variables, and in many of these cases, arithmetic averages are used. Therefore, in this study, we extended the arithmetic mean and considered the generalized weighted averages to balance the two variables. The generalized weighted averages for variables $x$ and $y$ are expressed in Equation (5).

$$\mu(x, y|\alpha, m) = [(1-\alpha)x^m + \alpha y^m]^{1/m}, \quad (5)$$

where $\alpha$ takes values in the range $0 \leq \alpha \leq 1$ and denotes weighting the values of $x$ and $y$, $m$ takes values in the range $-\infty \leq m \leq \infty$ and denotes the manner of taking the mean, and $\mu$ represents the average value calculated by the generalized weighting calculation. If $\alpha = 0.5$ and $m = 1.0$, then $\mu(x, y|0.5, 1.0) = (x + y)/2$ represents the arithmetic mean. If $\alpha = 0.5$ and $m = -1.0$, then $\mu(x, y|0.5, -1.0) = 2xy/(x + y)$ represents the harmonic mean. For $m = 0.0$, by taking the limit $m \to 0$, $k^m = e^{m \log k} \approx 1 + m \log k$ by the Maclaurin expansion, Equation (5) can be transformed into Equation (6).

$$\begin{aligned}
\mu(x, y|\alpha, m) &= [(1-\alpha)x^m + \alpha y^m]^{1/m} \\
&\approx [(1-\alpha)(1 + m \log x) + \alpha(1 + m \log y)]^{1/m} \\
&= [1 + m \log x^{1-\alpha} + m \log y^\alpha]^{1/m} \quad (6) \\
&\approx [(x^{1-\alpha} y^\alpha)^m]^{1/m} \\
&= x^{1-\alpha} y^\alpha
\end{aligned}$$

Thus, for $\alpha = 0.5$ and $m = 0.0$, $\mu(x, y|0.5, 0.0) = \sqrt{xy}$ represents the geometric mean. Figure 1 shows an overview of the generalized weighted averages.

By applying Equation (5) for generalized weighted averages to Equation (1) of UCB1, the UCB score is expanded as shown in Equation (7).

$$UCB_i^{GWA-UCB1} = \left[(1-\alpha)\bar{X}_i^m + \alpha \sqrt{\frac{2 \log n}{T_i(n)}}^m\right]^{1/m} \quad (7)$$

If $\alpha = 0.5$ and $m = 1.0$, Equation (7) is consistent with Equation (1) of UCB1. We call this the UCB1 algorithm that uses the UCB score in Equation (7), or GWA-UCB1. We explicitly call this algorithm GWA-UCB1 rather than GWA-UCB because the idea can be applied to other algorithms such as UCB1-Tuned. However, in this study, we limited the application of generalized weighted averages to UCB1 to validate this idea.

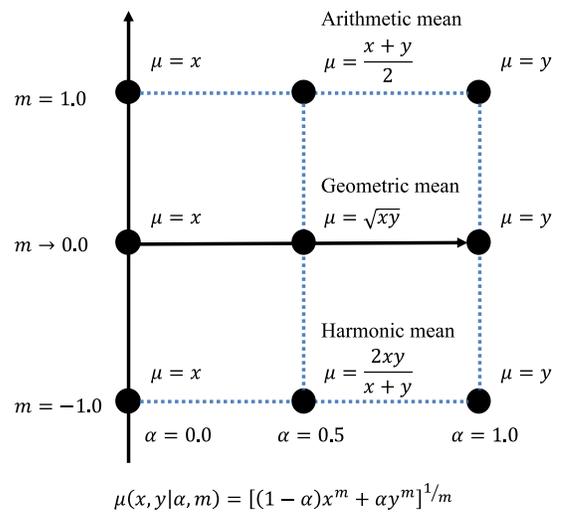

Figure 1: Overview of generalized weighted averages.

# Experiments

Preliminary experiments were first conducted on the stochastic MAB problem with $k = 2$ arms to determine a set of parameters for GWA-UCB1 that is valid in many problem settings. The optimal parameters for G-UCB1 were investigated to compare algorithms. Three experiments were conducted using the set of parameters obtained from preliminary experiments. The comparison algorithms for these three experiments used UCB1, UCB1-Tuned, and Thompson sampling, in addition to the parameter sets G-UCB1 and GWA-UCB1 obtained in the preliminary experiments. The detailed setup for each experiment is as follows.

## Preliminary Experiment

In a preliminary experiment, the optimal parameters for G-UCB1 and GWA-UCB1 were investigated by conducting 1,000 trials with 10,000 steps for a stochastic MAB problem with $k = 2$ arms. For G-UCB1, the value of parameter c in Equation (4) was shifted by a 0.01 increment in the interval $[0.05, 0.95]$, and the average regret was calculated for each parameter. For GWA-UCB1, values of parameters $\alpha$ and m in Equation (7) were each shifted by a 0.01 increment in the intervals $[0.05, 0.95]$ and $[-2.00, 4.00]$, and the average regret was calculated for each pair of parameters.

## Experiment 1

In Experiment 1, the average regret was calculated after 100,000 simulation trials with 10,000 steps each for the stochastic MAB problem with arms $k = 2, 8$, and 32. The reward probability $p_i$ for each arm was determined by a uniform distribution in the [0,1] interval for each trial. This experiment is the most commonly used problem setup to verify the algorithm's performance in a stochastic MAB problem, which confirms the standard performance of the proposed method.

## Experiment 2

In Experiment 1, we confirmed the performance of the algorithm with a small number of arms. However, when the number of arms is very large, it is likely that a few arms have high reward probabilities in actual situations. For example, for a Go player, one can see that although there are many options in a given scene, there are only a few options that lead to victory.

Therefore, in Experiment 2, the average regret was calculated for 10,000 trials of the simulation with 50,000 steps each for the number of arms $k = 32, 128$, and 512. The reward probability $p_i$ for each arm was determined by a normal distribution with a mean of 0.5 and a standard deviation of 0.1 for each trial. This reproduced an environment in which the reward probability for most arms was not high, around 0.5, and only some arms had high reward probability values.

## Experiment 3

In Experiment 3, the survival rate and average budget were calculated after 10,000 trials of 50,000 step simulations for the survival MAB problem for $k = 8, 32$, and 128 arms. The reward probability $p_i$ for one arm was set to 0.55 and 0.45 for the rest. This implies that the budget will become negative if the single best arm with $p_i = 0.55$ cannot be searched early in the process, which puts the problem at high risk of ruin. The initial budgets for the number of arms $k =$

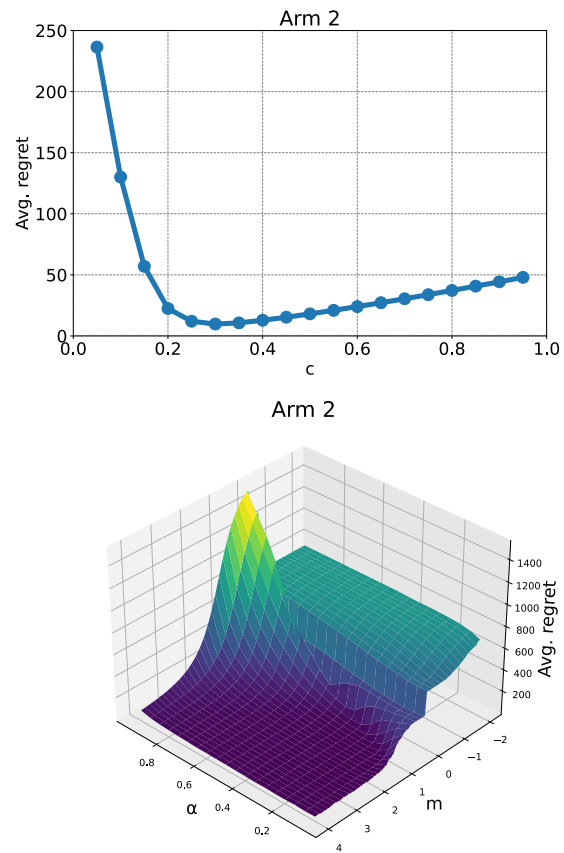

Figure 2: Preliminary experiment results. The upper figure shows the average regret of G-CUB1, and the lower figure shows the average regret of GWA-CUB1. The average regret is the value after 1,000 trials of a 10,000-step simulation for a stochastic MAB problem with $k = 2$ arms. The reward probability for each arm was determined by a uniform distribution on the [0,1] interval for each trial.

8, 32, and 128 were set to $b_0 = 80, 320,$ and $1{,}280$, respectively.

## Results

Preliminary experiments showed that the parameters that give the best results in the stochastic MAB problem for $k = 2$ arms were $c = 0.30$ for G-UCB1 and $\alpha = 0.21$ and $m = 1.30$ for GWA-UCB1. Figure 2 shows the results of the preliminary experiments and confirms the approximate results for the G-UCB1 and GWA-UCB1 parameters.

Figure 3 shows the results of Experiment 1. For $k = 2$ and $k = 8$ arms, GWA-UCB1 exhibited the best performance. For $k = 32$ arms, Thompson sampling performed the best. Therefore, when the reward probabilities of the arms were determined from a uniform distribution, as the number of arms $k$ increased, the performance difference between Thompson sampling and GWA-UCB1 narrowed, and Thompson sampling outperformed other algorithms. However, for a small number of arms $k$, at least $k \leq 8$, GWA-UCB1 exhibited excellent performance. Therefore, GWA-UCB1 can be effectively used to make the best choice from a small number of options, such as A/B testing, which is performed in internet marketing.

Figure 4 shows the results of Experiment 2. For all cases with $k = 32, 128,$ and $512$ arms, GWA-UCB1 exhibited the best performance. In other words, when the reward probabilities of the arms were determined by a normal distribution with mean 0.5 and standard deviation 0.1, GWA-UCB1 showed superior performance regardless of the number of arms $k$. Therefore, we believe that GWA-UCB1 can be effectively used in reinforcement learning tasks, such as Go and video games, where the number of correct answer options is small despite the large number of actions that can be selected.

Figure 5 shows the survival rate results for Experiment 3, and Figure 6 shows the average budget results in Experiment 3. In all cases, with $k = 8, 32,$ and $128$ arms, GWA-UCB1 exhibited the best performance. It is worth noting that even Thompson sampling, which showed excellent performance in the stochastic MAB problem, had a low survival rate in a very difficult setting, where the reward probability for 127 arms was 0.45 and only one arm had a 0.55 reward probability.

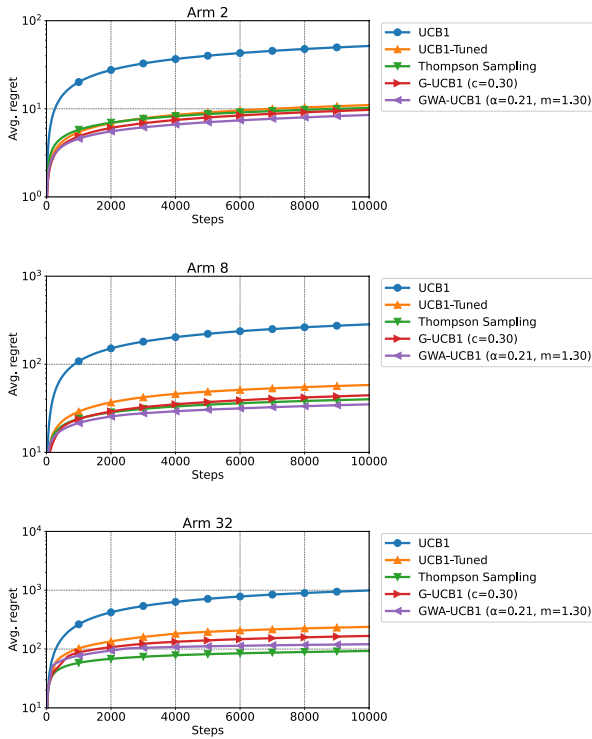

Figure 3: Experiment 1 results. The figure shows the average regret when the number of arms is $k = 2, 8,$ and 32, from top to bottom, respectively. The average regret is the value after 100,000 trials of a 10,000-step simulation for a stochastic MAB problem. The reward probability for each arm was determined by a uniform distribution on the [0,1] interval for each trial.

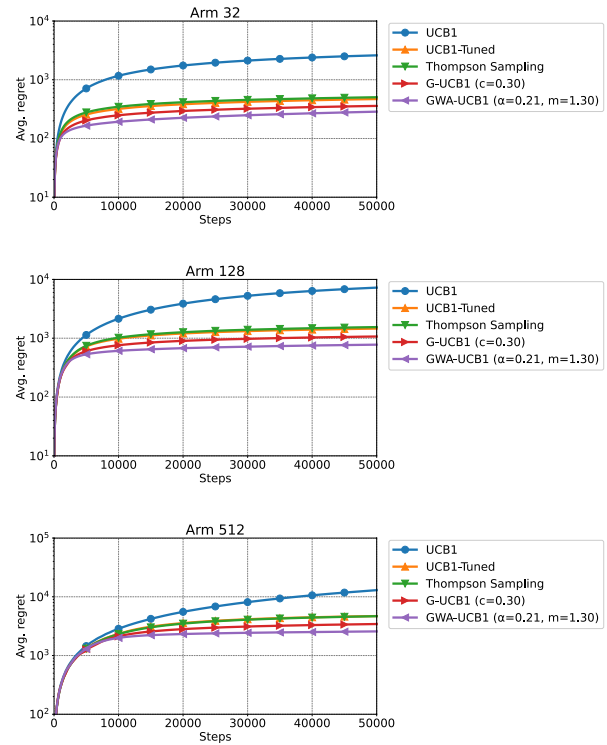

Figure 4: Experiment 2 results. The figure shows the average regret when the number of arms is $k = 32, 128,$ and 512, from top to bottom, respectively. The average regret is the value after 10,000 trials of a 50,000-step simulation for a stochastic MAB problem. The reward probability for each arm was determined by a normal distribution with a mean of 0.5 and a standard deviation of 0.1 for each trial.

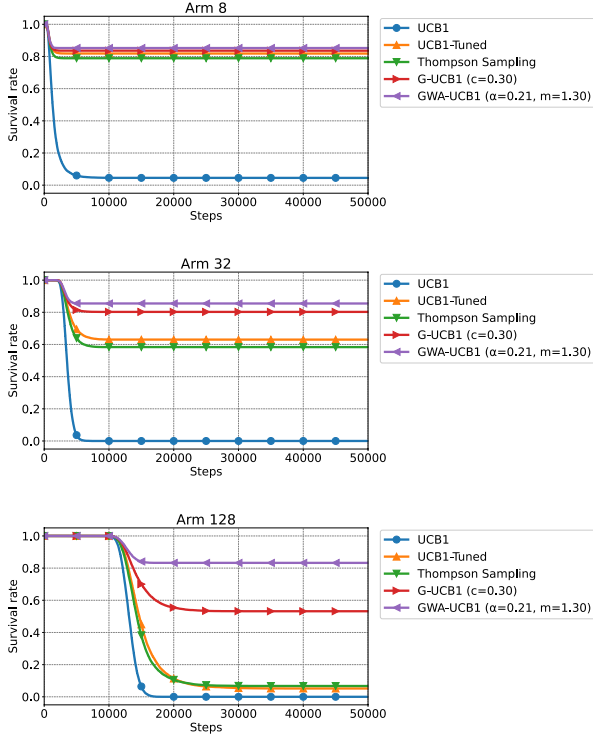
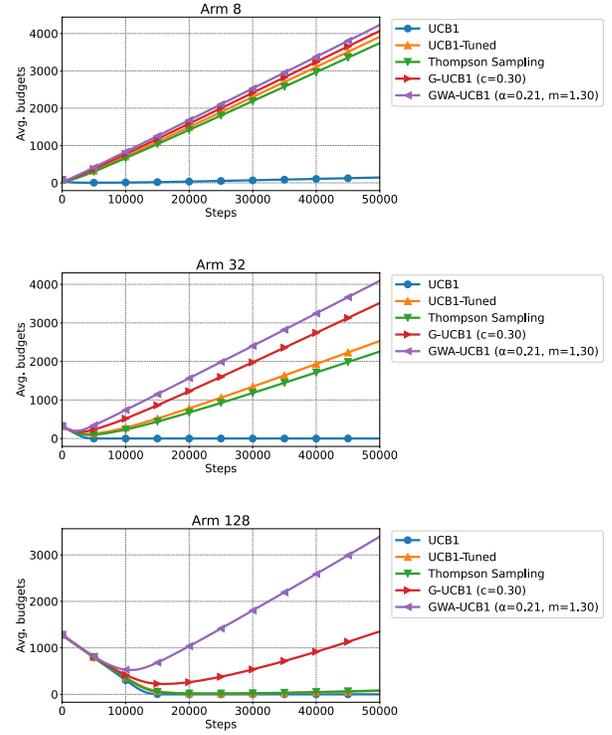

Figure 5: Experiment 3 results. The figure shows the survival rate when the number of arms is $k = 8$, 32, and 128, from top to bottom, respectively. The survival rate is the value after 10,000 trials of a 50,000-step simulation for a survival MAB problem. The reward probability for each arm was 0.55 for only one arm and 0.45 for the remaining arms.

Figure 6: Experiment 3 results. The figure shows the average budgets when the number of arms is $k = 8$, 32, and 128, from top to bottom, respectively. The average budget is the value after 10,000 trials of a 50,000-step simulation for a survival MAB problem. The reward probability for each arm was 0.55 for only one arm and 0.45 for the remaining arms.

By contrast, GWA-UCB1 maximized the budget more than the other algorithms while maintaining a high survival rate. Therefore, GWA-UCB1 can be effectively used in situations such as stock trading and casino gambling, where the budget is limited.

## Discussion

The practical parameters for GWA-UCB1 investigated in the preliminary experiments were $\alpha = 0.21$ and $m = 1.30$. Parameter $\alpha$ is a parameter that adjusts the weights of exploration and exploitation and has a similar role as parameter $c$ in G-UCB1. Therefore, considering the value $\alpha = 0.21$, it is more practical to emphasize exploitation over exploration in a stochastic MAB problem with $k = 2$ arms.

Parameter $m$ represents the manner of taking the mean. For clarity, let us consider Equation (5) with $\alpha = 0.5$ and $y = 1 - x (0.0 \leq x \leq 1.0)$. When $m > 1.0$, $\mu(x, 1-x|0.5, m)$ is a convex function, and the values at both ends, $\mu(0.0, 1.0|0.5, m)$ and $(1.0, 0.0|0.5, m)$, approach 1 as $m$ increases. When $m = 1.0$, $\mu(x, 1-x|0.5, 1.0) = 0.5$ regardless of the value of $x$. When $0.0 < m < 1.0$, $\mu(x, 1-x|0.5, m)$ is a concave function and the values at both ends, $\mu(0.0, 1.0|0.5, m)$ and $(1.0, 0.0|0.5, m)$, approach 0 as $m$ decreases. When $m \leq 0.0$, $\mu(x, 1-x|0.5, m)$ approaches 0 as the value of $x$ approaches 0 or 1. Figure 7 shows the values of generalized weighted averages when $\alpha = 0.5$ and $y = 1 - x$. However, interpreting how the value $m = 1.30$ contributes to the performance of GWA-UCB1 is difficult; hence, it is a subject for future work.

GWA-UCB1 showed superior performance not only in reward acquisition but also in terms of survival rate. There-fore, this heuristic, which is simple and achieves good performance with a small amount of computation, provides important perspectives in the research field of machine learning. Shinohara et al. measured how humans derive the strength of causal relationships between events from cause and effect using generalized weighted averages (Shinohara et al. 2020). This result achieves a better fit than previous studies, which also suggests the usefulness of balancing the two variables using generalized weighted averages.

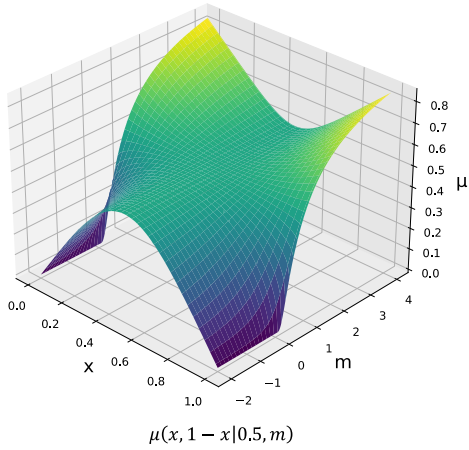

Figure 7: Examples of $\mu$ values of generalized weighted averages with $\alpha = 0.5$, $y = 1 - x$. Values of parameters $x$ and $m$ in Equation (5) were each shifted by 0.01 increment in the intervals $[0.00, 1.00]$ and $[-2.00, 4.00]$, and the $\mu$ value was calculated for each pair of parameters.

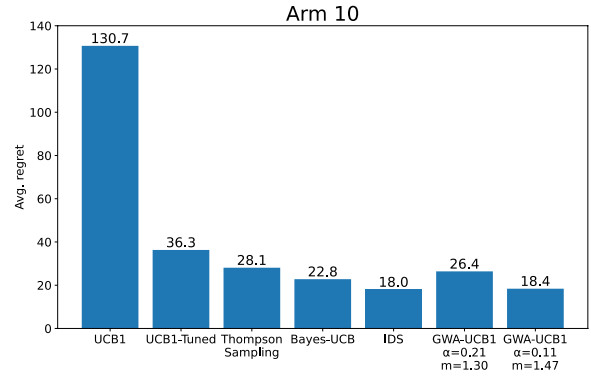

Figure 8: Comparative results of Russo and Van Roy's study (Russo and Van Roy 2018) and GWA-UCB1. The average regret is the value after 2,000 trials of a 1,000-step simulation for a stochastic MAB problem with $k = 10$ arms. The reward probability for each arm was determined by a uniform distribution on the [0,1] interval for each trial.

The optimal GWA-UCB1 parameters for the stochastic MAB problem with $k = 2$ have been investigated herein, but the parameters could also be adjusted according to the specific environment. The case with $k = 10$ arms was investigated in the same manner as in the preliminary experiment. As a result, $\alpha = 0.11$ and $m = 1.47$ were derived. We then replicated the experiment of Russo and Van Roy (Russo and Van Roy 2018) and compared the results of their study with those of the parameter-adjusted GWA-UCB1. Figure 8 presents the comparison results between Russo and Van Roy's study and the GWA-UCB1 results for a stochastic MAB problem with $k = 10$ arms. The results show that GWA-UCB1 with $\alpha = 0.11$ and $m = 1.47$ achieved a slightly worse performance than IDS, but as close as possible. By contrast, even GWA-UCB1 with $\alpha = 0.21$ and $m = 1.30$ achieved better performance than Thompson sampling and can be used effectively enough.

GWA-UCB1 can be implemented by simply replacing Equation (1) of UCB1 with Equation (7) and requires as little computation time as UCB1. Furthermore, because this is a simple modification, it is easy to implement and can be applied to various UCB-based reinforcement learning models. This study only shows empirical results for GWA-UCB1; therefore, a theoretical analysis should be conducted in the future.

## Conclusion

In this study, we extended UCB1 using generalized weighted averages and proposed a new generalized UCB called GWA-UCB1, which can be performed with less computation cost. Furthermore, GWA-UCB1 is a simple modification of the UCB1 formula that is easy to implement and can be extended to a variety of UCB-based reinforcement learning models. The results showed that GWA-UCB1 with $\alpha = 0.21$ and $m = 1.30$ achieved better performance than G-UCB1, UCB1-Tuned, and Thompson sampling in most problem settings and may be used in many situations.

Three main challenges must be addressed in the future: (1) Undertaking of a theoretical analysis concerning GWA-UCB1, (2) integration of GWA-UCB1 into UCB-based reinforcement learning models, coupled with the validation of its performance, and (3) application of generalized weighted averages to other machine learning tasks that must consider the trade-offs between the two variables in consideration. In the MAB problem, for example, to properly estimate the reward probability of an arm, the reward of that arm must be observed over a prolonged period. However, if an arm's reward probability is nonstationary, it must be evaluated over a short period. By applying generalized weighted averages to address such a dilemma, it may be possible to construct an MAB algorithm that is effective even in nonstationary environments.

# References

Auer, P.; Cesa-Bianchi, N.; and Fischer, P. 2002. Finite-time analysis of the multiarmed bandit problem. *Machine learning*, 47: 235–256.

Auer, P.; Jaksch, T.; and Ortner, R. 2008. Near-optimal regret bounds for reinforcement learning. Advances in neural information processing systems, 21.

Baek, J. and Farias, V. 2023. TS-UCB: Improving on Thompson sampling with little to no additional computation. In International Conference on Artificial Intelligence and Statistics (pp. 11132-11148). PMLR.

Bubeck, S. and Liu, C. Y. 2013. Prior-free and prior-dependent regret bounds for thompson sampling. Advances in neural information processing systems, 26.

Cappé, O.; Garivier, A.; Maillard, O. A.; Munos, R.; and Stoltz, G. 2013. Kullback-Leibler upper confidence bounds for optimal sequential allocation. *The Annals of Statistics*, 1516–1541.

Chapelle, O. and Li, L. 2011. An empirical evaluation of thompson sampling. Advances in neural information processing systems, 24.

Elena, G.; Milos, K.; and Eugene, I. 2021. Survey of multi-armed bandit algorithms applied to recommendation systems. *International Journal of Open Information Technologies*, 9(4): 12–27.

Garivier, A. and Cappé, O. 2011. The KL-UCB algorithm for bounded stochastic bandits and beyond. In Proceedings of the 24th annual conference on learning theory (pp. 359-376). JMLR Workshop and Conference Proceedings.

Garivier, A.; Lattimore, T.; and Kaufmann, E. 2016. On explore-then-commit strategies. Advances in Neural Information Processing Systems, 29.

Jin, C.; Allen-Zhu, Z.; Bubeck, S.; and Jordan, M. I. 2018. Is Q-learning provably efficient? Advances in neural information processing systems, 31.

Kaufmann, E.; Cappé, O.; & Garivier, A. 2012a. On Bayesian upper confidence bounds for bandit problems. In Artificial intelligence and statistics (pp. 592-600). PMLR.

Kaufmann, E.; Korda, N.; and Munos, R. 2012b. Thompson sampling: An asymptotically optimal finite-time analysis. In International conference on algorithmic learning theory (pp. 199-213). Berlin, Heidelberg: Springer Berlin Heidelberg.

Kaufmann, E. 2018. On Bayesian index policies for sequential resource allocation. *The Annals of Statistics*, 46(2): 842–865.

Kocsis, L. and Szepesvári, C. 2006. Bandit based monte-carlo planning. In European conference on machine learning (pp. 282-293). Berlin, Heidelberg: Springer Berlin Heidelberg.

Korkut, M. and Li, A. 2021. Disposable Linear Bandits for Online Recommendations. *Proceedings of the AAAI Conference on Artificial Intelligence*, 35(5): 4172–4180.

Liu, G.; Shi, W.; and Zhang, K. 2019. An upper confidence bound approach to estimating coherent risk measures. In 2019 Winter Simulation Conference (WSC) (pp. 914-925). IEEE.

Ménard, P. and Garivier, A. 2017. A minimax and asymptotically optimal algorithm for stochastic bandits. In International Conference on Algorithmic Learning Theory (pp. 223-237). PMLR.

Moraes, R.; Marino, J.; Lelis, L.; and Nascimento, M. 2018. Action abstractions for combinatorial multi-armed bandit tree search. *In Proceedings of the AAAI Conference on Artificial Intelligence and Interactive Digital Entertainment*, 14 (1): 74–80.

Nuara, A.; Trovo, F.; Gatti, N.; and Restelli, M. 2018. A combinatorial-bandit algorithm for the online joint bid/budget optimization of pay-per-click advertising campaigns. *Proceedings of the AAAI Conference on Artificial Intelligence*, 32(1).

Perotto, F. S.; Bourgais, M.; Silva, B. C.; and Vercouter, L. (2019, June). Open problem: Risk of ruin in multiarmed bandits. In Conference on Learning Theory (pp. 3194-3197). PMLR.

Osband, I.; Russo, D.; and Van Roy, B. 2013. (More) efficient reinforcement learning via posterior sampling. Advances in Neural Information Processing Systems, 26.

Riou, C.; Honda, J.; and Sugiyama, M. 2022. The Survival Bandit Problem. arXiv preprint arXiv:2206.03019.

Robbins, H. 1952. Some aspects of the sequential design of experiments. *Bulletin of the American Mathematical Society*, 58(5): 527–535.

Russo, D. and Van Roy, B. 2016. An information-theoretic analysis of thompson sampling. *The Journal of Machine Learning Research*, 17(1): 2442–2471.

Russo, D. and Van Roy, B. 2018. Learning to optimize via information-directed sampling. *Operations Research*, 66(1): 230–252.

Schrittwieser, J.; Antonoglou, I.; Hubert, T.; Simonyan, K.; Sifre, L.; Schmitt, S.; ... and Silver, D. 2020. Mastering atari, go, chess and shogi by planning with a learned model. *Nature*, 588(7839): 604–609.

Schwartz, E. M.; Bradlow, E. T.; and Fader, P. S. 2017. Customer acquisition via display advertising using multi-armed bandit experiments. *Marketing Science*, 36(4): 500–522.

Shinohara, S.; Manome, N.; Suzuki, K.; Chung, U. I.; Takahashi, T.; Gunji, P. Y.; ... & Mitsuyoshi, S. 2020. Extended Bayesian inference incorporating symmetry bias. *Biosystems*, 190: 104104.

Silva, N.; Werneck, H.; Silva, T.; Pereira, A. C.; and Rocha, L. 2022. Multi-armed bandits in recommendation systems: A survey of the state-of-the-art and future directions. *Expert Systems with Applications*, 197: 116669.

Silver, D.; Schrittwieser, J.; Simonyan, K.; Antonoglou, I.; Huang, A.; Guez, A.; ... and Hassabis, D. 2017. Mastering the game of go without human knowledge. *Nature*, 550(7676): 354–359.

Silver, D.; Hubert, T.; Schrittwieser, J.; Antonoglou, I.; Lai, M.; Guez, A.; ... & Hassabis, D. 2018. A general reinforcement learning algorithm that masters chess, shogi, and Go through self-play. *Science*, 362(6419): 1140–1144.

Sutton, R. S. and Barto, A. G. 1998. *Introduction to reinforcement learning*, 135: 223–260. Cambridge: MIT press.

Świechowski, M.; Godlewski, K.; Sawicki, B.; and Mańdziuk, J. 2023. Monte Carlo tree search: A review of recent modifications and applications. *Artificial Intelligence Review*, 56(3): 2497–2562.

Thompson, W. R. 1933. On the likelihood that one unknown probability exceeds another in view of the evidence of two samples. *Biometrika*, 25(3-4): 285–294.

Tran-Thanh, L.; Chapman, A.; De Cote, E. M.; Rogers, A.; and Jennings, N. R. 2010. Epsilon–first policies for budget–limited multi-armed bandits. *Proceedings of the AAAI Conference on Artificial Intelligence*, 24 (1), 1211–1216.

Xu, M.; Qin, T.; and Liu, T. Y. 2013. Estimation bias in multi-armed bandit algorithms for search advertising. Advances in Neural Information Processing Systems, 26.